\documentclass{Interspeech2024}

\usepackage[table,dvipsnames]{xcolor}
\usepackage{todonotes}
\usepackage{amsmath,amssymb,graphicx}
\usepackage{tabularx}
\usepackage{multirow}
\usepackage{ctable,booktabs}
\usepackage[inline]{enumitem}
\usepackage{xspace}

\usepackage[
backend=biber,
style=ieee,
citestyle=numeric-comp,
maxbibnames=3,
maxcitenames=3,
doi=false,isbn=false,url=false,eprint=false
]{biblatex}
\defbibheading{bibliography}[\refname]{}

\DeclareSourcemap{
    \maps[datatype=bibtex, overwrite=true]{
        \map{
            \step[fieldsource=booktitle,
            match=\regexp{.*Interspeech.*},
            replace={Proc. Interspeech}]
            \step[fieldsource=journal,
            match=\regexp{.*INTERSPEECH.*},
            replace={Proc. Interspeech}]
            \step[fieldsource=booktitle,
            match=\regexp{.*ICASSP.*},
            replace={Proc. ICASSP}]
            \step[fieldsource=booktitle,
            match=\regexp{.*icassp_inpress.*},
            replace={Proc. ICASSP (in press)}]
            \step[fieldsource=booktitle,
            match=\regexp{.*Acoustics,.*Speech.*and.*Signal.*Processing.*},
            replace={Proc. ICASSP}]
            \step[fieldsource=booktitle,
            match=\regexp{.*International.*Conference.*on.*Learning.*Representations.*},
            replace={Proc. ICLR}]
            \step[fieldsource=booktitle,
            match=\regexp{.*International.*Conference.*on.*Computational.*Linguistics.*},
            replace={Proc. COLING}]
            \step[fieldsource=booktitle,
            match=\regexp{.*SIGdial.*Meeting.*on.*Discourse.*and.*Dialogue.*},
            replace={Proc. SIGDIAL}]
            \step[fieldsource=booktitle,
            match=\regexp{.*International.*Conference.*on.*Machine.*Learning.*},
            replace={Proc. ICML}]
            \step[fieldsource=booktitle,
            match=\regexp{.*North.*American.*Chapter.*of.*the.*Association.*for.*Computational.*Linguistics:.*Human.*Language.*Technologies.*},
            replace={Proc. NAACL}]
            \step[fieldsource=booktitle,
            match=\regexp{.*Empirical.*Methods.*in.*Natural.*Language.*Processing.*},
            replace={Proc. EMNLP}]
            \step[fieldsource=booktitle,
            match=\regexp{.*Association.*for.*Computational.*Linguistics.*},
            replace={Proc. ACL}]
            \step[fieldsource=booktitle,
            match=\regexp{.*Automatic.*Speech.*Recognition.*and.*Understanding.*},
            replace={Proc. ASRU}]
            \step[fieldsource=booktitle,
            match=\regexp{.*Spoken.*Language.*Technology.*},
            replace={Proc. SLT}]
            \step[fieldsource=booktitle,
            match=\regexp{.*Speech.*Synthesis.*Workshop.*},
            replace={Proc. SSW}]
            \step[fieldsource=booktitle,
            match=\regexp{.*workshop.*on.*speech.*synthesis.*},
            replace={Proc. SSW}]
            \step[fieldsource=booktitle,
            match=\regexp{.*Advances.*in.*neural.*information.*processing.*},
            replace={Proc. NeurIPS}]
            \step[fieldsource=booktitle,
            match=\regexp{.*Advances.*in.*Neural.*Information.*Processing.*},
            replace={Proc. NeurIPS}]
            \step[fieldsource=booktitle,
            match=\regexp{.*Workshop.*on.* Applications.* of.* Signal.*Processing.*to.*Audio.*and.*Acoustics.*},
            replace={Proc. WASPAA}]
            \step[fieldsource=publisher,
            match=\regexp{.+},
            replace={{}}]
            \step[fieldsource=month,
            match=\regexp{.+},
            replace={{}}]
            \step[fieldsource=location,
            match=\regexp{.+},
            replace={{}}]
            \step[fieldsource=address,
            match=\regexp{.+},
            replace={{}}]
            \step[fieldsource=organization,
            match=\regexp{.+},
        replace={{}}]
        }
    }
}

\newcommand{\dataset}{\textsc{Mcv-Accent}}

\interspeechcameraready

\title{Improving Self-supervised Pre-training using Accent-Specific Codebooks}

\name[affiliation={*1}]{Darshan}{Prabhu}
\name[affiliation={*1}]{Abhishek}{Gupta}
\name[affiliation={1}]{Omkar}{Nitsure}
\name[affiliation={1}]{Preethi}{Jyothi}
\name[affiliation={2}]{Sriram}{Ganapathy}

\address{
$^1$Indian Institute of Technology Bombay, Mumbai, India \\
$^2$Indian Institute of Science, Bangalore, India.
}
\email{\{darshanp, abhishekgupta, pjyothi\}@cse.iitb.ac.in, sriramg@iisc.ac.in}


\begin{document}

\maketitle

\begin{abstract}
    
    Speech accents present a serious challenge to the performance of state-of-the-art end-to-end Automatic Speech Recognition (ASR) systems. Even with self-supervised learning and pre-training of ASR models, accent invariance is seldom achieved. In this work, we propose an accent-aware adaptation technique for self-supervised learning that introduces a trainable set of accent-specific codebooks to the self-supervised architecture. These learnable codebooks enable the model to capture accent specific information during pre-training, that is further refined during ASR finetuning. On the Mozilla Common Voice dataset, our proposed approach outperforms all other accent-adaptation approaches on both seen and unseen English accents, with up to $9\%$ relative reduction in word error rate (WER). 
\looseness=-1
\end{abstract}

\def\thefootnote{*}\footnotetext{These authors contributed equally to this work.}\def\thefootnote{\arabic{footnote}}

\section{Introduction}
Self-supervised learning (SSL) has been established as a powerful technique to learn representations for speech and language processing~\cite{layerwise_analysis}. With pretrained SSL models as a starting point, even small amounts of labeled data are sufficient to achieve commercially acceptable results in various downstream speech tasks~\cite{baevski2020wav2vec, zhao2022improving}. This has led to the wide adoption of the current defacto training regime of pretraining a model with an SSL objective, followed by fine-tuning on downstream tasks such as automatic speech recognition (ASR) and spoken language understanding~\cite{yang21c_interspeech, tsai-etal-2022-superb}. A key limitation of SSL-based models is that performance on the downstream task suffers if there is a domain shift from the pretraining data~\cite{hsu2021robust}. In this work, we focus on speech accents as our domain of interest. The problem statement can be outlined as - \emph{How can we make SSL-trained models more robust to varying speech accents, both seen and unseen during training?} 

While prior work has extensively explored improving accented ASR in the fine-tuning stage~\cite{asr_residual, jain2018improved, winata2020learning, accentgrapheme, layer_accnt, e2emultitask}, limited work has gone into improving the robustness of SSL pretraining to varying speech accents. Prior work on accent adaptation of SSL models includes introducing an additional classifier into SSL pretraining~\cite{deng21b_interspeech}, employing accent-specific adapters~\cite{Bhatia2023}, and normalizing pseudo-targets to a single accent~\cite{Poncelet2023UnsupervisedAA}. 

In this work, we adapt the technique introduced in Prabhu et al.~\cite{Prabhu2023AccentedSR} for SSL pretraining. We introduce accent information during self-supervised pre-training via a set of accent-specific codebooks. These codebooks are integrated into the model using a cross-attention module. For each accent seen in the training data, we introduce a learnable codebook consisting of vectors that capture accent-specific information during the pre-training process. While \cite{Prabhu2023AccentedSR} used this technique exclusively during ASR finetuning, we show that the  benefits increase when used during the SSL stage. To evaluate the effectiveness of our method, we show ASR experiments on the multi-accented Mozilla Common Voice corpus and show significant word error rate (WER) reductions compared to multiple variants of HuBERT and other accent adaptation techniques including an accent-agnostic approach (Domain Adversarial Training~(DAT)~\cite{bobw}) and an accent-aware approach (MultiTask Learning~(MTL)~\cite{asr_clf}). 

While many self-supervised models exist for speech-related tasks~\cite{baevski2020wav2vec, chen2022wavlm, chung2021w2v, discretebert, decoar2, vqwav2vec, hubert}, in this work we use HuBERT --- a state-of-the-art architecture that uses a masked language modeling (MLM) objective and aims to reconstruct a sequence of discrete pseudo-targets derived via a $k$-means clustering step. The clustering step in HuBERT is used to generate pseudo-targets for pretraining. Prior work has aimed to improve the quality of these generated pseudo-targets either through teacher-forcing~\cite{hubert_teacher} or by designing task-specific pseudo-targets~\cite{ssldeepcluster}. Other efforts have focused on improving the HuBERT architecture itself~\cite{fan2023ctcbert, jointssldecoder}. To the best of our knowledge, adapting HuBERT pretraining to be more robust across accents has not been attempted before.

In summary, our main contributions are as follows:
\begin{itemize}
    \item We introduce codebook-based accent adaptation for self-supervised pre-training that aims at using a collection of accent-specific codebooks and cross-attention to improve SSL representations in the presence of accents.%
    \footnote{Code is available at \href{https://github.com/csalt-research/accented-codebooks-asr/tree/accented-pretraining}{https://github.com/csalt-research/accented-codebooks-asr/tree/accented-pretraining}.}
    Our model outperforms baselines and all other previous approaches and achieves a significant improvement in performance with up to 9\% relative WER reduction on the MCV dataset.
    \item In a zero-shot setting on the L2-Arctic dataset, we show that our codebook-based pretrained model significantly outperforms baselines, thus demonstrating the generalization capability of our proposed system.
\end{itemize}


\section{Methodology}


\begin{figure*}[t]
  \centering
  \includegraphics[keepaspectratio, width=0.85\textwidth]{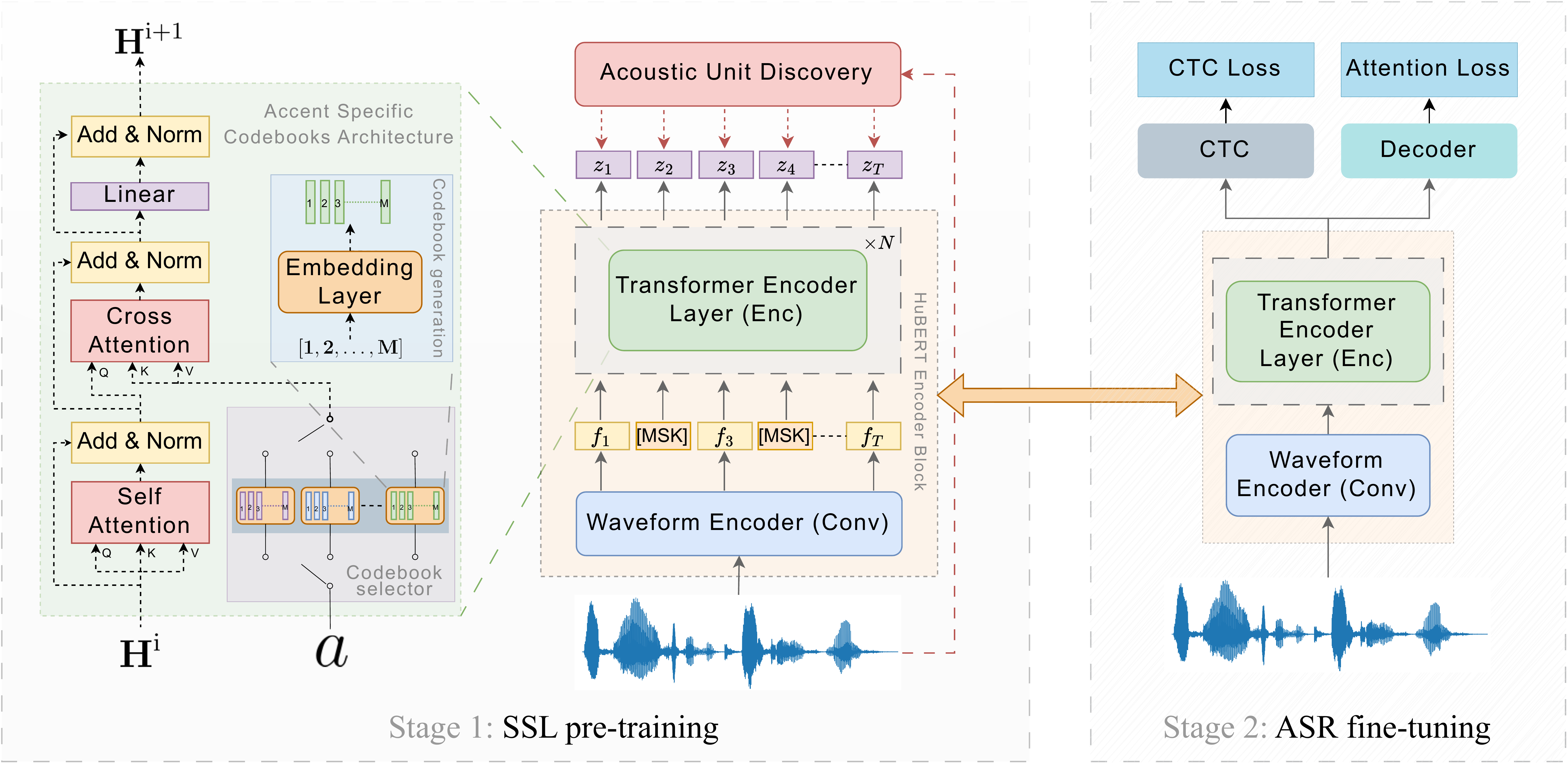}
  
  \caption{Overview of the two-stage training pipeline used in our proposed approach. In stage 1, we train an Encoder only model with the HuBERT style pre-training objective. Our architecture incorporates Accent-specific codebooks and additional Cross-Attention layers. The model trained in this stage is used as Encoder in stage 2, where we train an Encoder-Decoder model with Joint CTC-Attention objective for ASR fine-tuning. }   
  \label{fig:schematic} 
 
\vspace{-0.15in} 
\end{figure*}
Figure~\ref{fig:schematic} illustrates the main workflow of our proposed technique. We adopt the standard two-stage training pipeline used in self-supervised architectures: 

\begin{enumerate}
\item A \emph{self-supervised pretraining} stage that involves training an encoder using a self-supervised learning objective~\cite{mohamed2022self}. 
\item A \emph{supervised ASR fine-tuning} stage that uses the pretrained encoder from the previous stage and further finetunes it using an ASR-specific supervised loss.
\end{enumerate}

In our proposed framework, both these stages benefit from accent-specific codebooks elaborated in Sections $\S$\ref{subsec:method:pretrain} and $\S$\ref{subsec:method:asrfinetune}, respectively.

\subsection{Self-supervised Pre-training with Codebooks}\label{subsec:method:pretrain}

We use the HuBERT self-supervised architecture~\cite{hubert} that consists of three modules: a convolution-based waveform encoder~(denoted by \textsc{Conv}), a Transformer-based encoder~(denoted by $\textsc{Enc}$) and a projection~($\textsc{FFN}_{\text{tok}}$) module. \textsc{Conv} takes raw speech $X=\{ \mathbf{x}_1, \mathbf{x}_2, \ldots, \mathbf{x}_{L} | \mathbf{x}_i \in \mathbb{R} \}$ as its input, and maps it via a stack of convolutions to $F = \textsc{Conv}(X) = \{ \mathbf{f}_1, \mathbf{f}_2, \ldots, \mathbf{f}_{T} | \mathbf{f}_i \in \mathbb{R}^{d} \} $. 
$F$ is subsequently masked to generate $\tilde{F}$, where a fixed $P\%$ of its frames are randomly masked. This masked representation $\tilde{F}$ further passes through $\textsc{Enc}$ to generate a contextualized representation $H=\textsc{Enc}(\tilde{X}) = \{ \mathbf{h}_1, \mathbf{h}_2, \ldots \mathbf{h}_T | \mathbf{h}_i \in \mathbb{R}^{d} \} $. In parallel, an Acoustic Unit Discovery~(AUD) module directly acts on $X$ as its input and generates pseudo-target labels $Z=\{\mathbf{z}_1,\mathbf{z}_2,\ldots, \mathbf{z}_T | \mathbf{z}_i \in [1,V] \}$ via an offline clustering step that uses a single or ensemble of $k$-means clusterings. The size of the label set $V$ is determined by the number of clusters used in AUD. A simple projection layer $\textsc{FFN}_{\text{tok}}$ is used to convert $H$ from $\mathbb{R}^d$ to the target vocabulary $\mathbb{R}^V$, and the HuBERT model is trained end-to-end to predict $Z$ using a weighted cross-entropy loss. 
\looseness=-1
\begin{table*}[t!]
\caption{Comparison of the performance~(WER \% ) of our architecture~(\textsc{Codebook} Attention applied at all layers with $50$ entries in each learnable codebook) with baseline and other architectures on the \dataset-100 dataset. Numbers in bold denote the best across baselines, and \colorbox{green!20}{$\phantom{x}$} denotes the best WER across all experiments. Ties are broken using overall WER. The aggregate result for proposed model achieves statistically significant improvements compared to best baseline~(at $p<$ 0.001 MAPSSWE  test \cite{mapsswe}).}
\label{table:main_res}
\centering
\resizebox{\linewidth}{!}{
    \begin{tabular}{ l|c|c|cc|ccccc|ccccccccccc }
        \hline
        \hline 
        
        \multicolumn{1}{c|}{\multirow{2}{*}{\textbf{\small{Method}}}} & \multicolumn{1}{c|}{\multirow{2}{*}{\textbf{\small{Size}}}} & \multicolumn{3}{c|}{\textbf{\small{Aggregated}}} & \multicolumn{5}{c|}{\textbf{\small{Seen Accents}}} & \multicolumn{9}{c}{\textbf{\small{Unseen Accents}}} \\
        
        \cline{3-19}

        & & \textbf{\scriptsize{All}} & \textbf{\scriptsize{Seen}} & \textbf{\scriptsize{Unseen}} & \textbf{\scriptsize{AUS}} & \textbf{\scriptsize{CAN}} & \textbf{\scriptsize{UK}} &
        \textbf{\scriptsize{SCT}}  & \textbf{\scriptsize{US}}  & 
        
        \textbf{\scriptsize{AFR}} & \textbf{\scriptsize{HKG}} & \textbf{\scriptsize{IND}} & \textbf{\scriptsize{IRL}} & \textbf{\scriptsize{MAL}} & \textbf{\scriptsize{NWZ}} & \textbf{\scriptsize{PHL}} & \textbf{\scriptsize{SGP}} & \textbf{\scriptsize{WLS}}  \\
        
        \hline

        \small{Conformer~\cite{conformer}} & \small{43M} & \small{18.9} &  \small{14.0} &  \small{23.7} &  \small{13.8} & \small{15.0} & \small{15.7} & \small{13.4} & \small{13.3} & \small{21.5} & \small{27.2} & \small{29.4} & \small{21.4} & \small{32.2} & \small{19.9} & \small{26.1} & \small{34.7} & \small{17.9} \\
        
        \small{HuBERT~\cite{hubert}} & \small{104M} & \small{13.1} &  \small{9.1} &  \small{17.1} &  \small{9.0} & \small{10.3} & \small{9.4} & \small{7.8} & \small{8.7} & \small{15.6} & \small{20.0} & \small{18.7} & \small{16.3} & \small{23.5} & \small{14.0} & \small{20.3} & \small{24.5} & \small{9.8} \\
    
        \small{ + Pretrain ckpt} & \small{104M} & \small{9.7} &  \small{6.3} &  \small{13.1} &  \small{5.3} & \small{7.7} & \small{6.2} & \small{4.7} & \small{6.1} & \small{12.4} & \small{15.5} & \small{13.8} & \small{12.5} & \small{18.8} & \small{10.0} & \small{15.2} & \small{20.0} & \small{7.8} \\

        \small{ + Frozen layers} & \small{74M} & \small{\textbf{9.3}} &  \small{\textbf{6.0}} &  \small{\textbf{12.5}} &  \small{\textbf{4.9}} & \small{7.8} & \small{\textbf{5.5}} & \small{4.7} & \small{\textbf{5.9}} & \small{\textbf{11.6}} & \small{15.8} & \small{\textbf{12.2}} & \small{\textbf{11.8}} & \small{18.0} & \small{\textbf{9.6}} & \small{\textbf{15.0}} & \small{19.8} & \small{\textbf{7.3}} \\
    
        \small{MTL~\cite{asr_clf}} & \small{74M} &\small{9.4} &  \small{6.0} & \small{12.8} & \small{5.0} & \small{7.6} & \small{5.6} & \small{4.7} & \small{5.9} & \small{11.9} & \small{15.5} & \small{13.4} & \small{12.2} & \small{17.6} & \small{10.0} & \small{15.4} & \small{\textbf{19.1}} & \small{8.3} \\
    
        \small{DAT}~\cite{bobw} & \small{74M} & \small{9.3} &  \small{6.0} & \small{12.5} &  \small{5.1} & \small{\textbf{7.5}} & \small{5.9} & \cellcolor{green!20} \small{\textbf{4.4}} & \small{5.9} & \small{11.6} & \small{\textbf{15.2}} & \small{12.3} & \small{12.3} & \small{\textbf{17.1}} & \small{9.6} & \small{15.4} & \small{19.4} & \small{7.9}\\
    
        \hline
    
        \small{\textsc{Proposed}} & \small{76M} & \cellcolor{green!20} \small{\textbf{8.9}} &  \cellcolor{green!20} \small{\textbf{5.9}} & \cellcolor{green!20} \small{\textbf{11.9}} &  \cellcolor{green!20} \small{ \textbf{3.7}} & \cellcolor{green!20} \small{ \textbf{7.5}} & \cellcolor{green!20} \small{ \textbf{5.5}} & \small{ \textbf{4.7}} & \cellcolor{green!20} \small{ \textbf{5.9}} & \cellcolor{green!20} \small{ \textbf{10.8}} & \cellcolor{green!20} \small{ \textbf{14.7}} & \cellcolor{green!20} \small{ \textbf{11.6}} & \cellcolor{green!20} \small{ \textbf{11.4}} & \cellcolor{green!20} \small{ \textbf{16.7}} & \cellcolor{green!20} \small{ \textbf{9.2}} & \cellcolor{green!20} \small{ \textbf{14.9}} & \cellcolor{green!20} \small{ \textbf{18.3}} & \cellcolor{green!20} \small{ \textbf{7.3}} \\
        
        \hline
        \hline
    \end{tabular}
}

\vspace{-0.15in}
\end{table*}
%

Our accent-based modifications are restricted to the Transformer-based encoder $\textsc{Enc}$. The $\textsc{Enc}$ module consists of a stack of $N$ identical encoder layers. Each layer attends to the output of the previous layer and contextualizes it using self-attention blocks and projection layers. The self-attention block is responsible for introducing global context, while the projection layer refines the point-wise information. In addition to this, we introduce a cross-attention block that utilizes attention mechanism to incorporate information from accent-specific codebooks into the representations. We first discuss how the codebooks are generated followed by an explanation of how they are utilized in a single encoder layer. 

\vspace{0.15cm}
\noindent \textbf{Codebook Setup.} Let $E$ be the number of accents seen during training. We define a set of accent-specific codebooks $C=\{ C^1, C^2, \ldots C^E | C^i \in \mathbb{R}^{M \times d} \}$. Each $C_i$ comprises $M$ learnable codebook entries. 
That is, each codebook $C^i$ is:
\begin{align}
    C^i = \{\mathbf{c}^i_1,\ldots, \mathbf{c}^i_{M}| \mathbf{c}^i_j \in \mathbb{R}^d\} = \texttt{Embedding}([1,2,\ldots,M]) \nonumber
\end{align}
where \texttt{Embedding} is a standard embedding layer. For a training speech input $X$, whose underlying accent ID (indexing the $E$ seen accents) is $a \in \{1,\ldots,E\}$, we deterministically select the codebook $C^a$. The codebook entries in $C^a$ are subsequently integrated with all the encoder layers and across all attention heads. We note here that the codebooks are accent-specific on account of a single codebook being deterministically chosen per utterance during training.

\vspace{0.15cm}
\noindent \textbf{Integrating Codebooks using Cross-attention.} Each Transformer-based encoder layer of the HuBERT architecture consists of a self-attention block that introduces global context, followed by a position-wise feed-forward block that refines point-wise information. These two blocks are separated by layer normalization and coupled via residual connections. In our method, we introduce a cross-attention module that is positioned between the self-attention and feedforward blocks and integrates information from the codebooks into the audio representations generated after self-attention. More precisely, for the $i^\text{th}$ encoder layer, let $\mathbf{A}$ be the input to the cross-attention block and $\hat{\mathbf{A}}$ be the output. Then, the attention probabilities across the $M$ codebook entries in $C^a$ for the $j^\text{th}$ position $\mathbf{A}_{j}$ is computed as: 
\looseness=-1
\begin{align*}
\{\beta^j_{1}, \beta^j_{2},\ldots,\beta^j_{M}\} = 
\mathrm{softmax}\left(\frac{(\mathbf{A}_j W^i_{Q})(C^a W^i_{K})^T}{\sqrt{d}} \right) \nonumber
\end{align*}
\noindent where $W^i_{Q}$, $W^i_{K} \in \mathbb{R}^{d\times d}$ are learned projection matrices of the cross-attention block of the $i^\text{th}$ encoder layer and $\beta^j_k \in [0,1]$ is the attention weight assigned by the $j^\text{th}$ representation to the $k^{th}$ entry in codebook $C^a$. Finally, the $j^{th}$ frame in the output $\hat{\mathbf{A}}$ is a weighted average of the entries in codebook $C^a$:
\vspace{-0.2cm}
\begin{align*}
    \mathbf{\hat{A}}_{j} = \sum_{k=1}^{M} \beta_{k}^j \cdot (C^a_{k} W^i_{V})
\end{align*}

\noindent where $\hat{\mathbf{A}}_{j}$, $C^a_{k}$ $\in \mathbb{R}^d$ and $W^i_{V} \in \mathbb{R}^{d\times d}$. The output of the cross-attention block is subjected to layer normalization, residual connections are added and is fed as input to the feed-forward block. 
\looseness=-1

\subsection{Supervised ASR Fine-tuning}\label{subsec:method:asrfinetune}
For the second stage of ASR fine-tuning, we adopt the state-of-the-art hybrid CTC-attention end-to-end ASR framework~\cite{jointctc} that consists of three modules: an encoder~($\textsc{Enc}_f$), a decoder~(\textsc{Dec}) and a Connectionist Temporal Classification~(\textsc{CTC})~\cite{ctc} module. We replace the encoder $\textsc{Enc}_{f}$ with the pretrained \textsc{Conv} and $\textsc{Enc}$ modules from Section~\ref{subsec:method:pretrain}, pretrained in conjunction with accent codebooks.  ASR fine-tuning makes use of labeled speech instances $(X, Y)$ where $X$ is a raw speech sequence $X=\{ \mathbf{x}_1, \mathbf{x}_2, \ldots, \mathbf{x}_{L} | \mathbf{x}_i \in \mathbb{R} \}$ and $Y$ is a token sequence $\{y_1,\ldots, y_{M}\} $.  Both the encoder parameters and the accent-specific codebooks in $\textsc{Enc}_{f}$, initialized from the pretrained model, are further finetuned using a supervised ASR objective. The \textsc{Dec} module uses a cross-entropy loss ($\mathcal{L}_{\text{att}}$) to autoregressively predict a token $y_t$ given previous tokens $\{y_1,\ldots, y_{t-1}\} $ and the encoder outputs. The \textsc{CTC} module, on the other hand, imposes a CTC loss ($\mathcal{L}_{\text{ctc}}$) directly on the encoder outputs and predicts a frame-aligned sequence of tokens by marginalizing over all possible alignments of $Y$ to the encoder outputs. The final loss is the weighted sum of both losses:
\looseness=-1
\begin{align}
      \mathcal{L}_{\text{asr}} &= \eta \times \mathcal{L}_{\text{ctc}} + (1-\eta) \times \mathcal{L}_{\text{att}}
\end{align}
where $\eta$ is a hyperparameter that balances the objectives. 

\subsection{Inference using Codebooks}\label{subsec:method:inference}
During inference, we do not assume access to an accent label for a test utterance. We employ the joint-beam search proposed by Prabhu et al.~\cite{Prabhu2023AccentedSR}. This technique involves performing beam search jointly over all the seen accents. Scores for the seen accents are computed using each underlying accent-specific codebook. The beam width holds the best expansions across all seen accents. More details of the joint beam algorithm are in~\cite{Prabhu2023AccentedSR}.

\section{Experimental Setup}

We conduct all our pre-training and fine-tuning experiments using Fairseq~\cite{fairseq} and ESPnet~\cite{espnet} toolkits, respectively, on NVIDIA RTX A6000 GPUs. 

\vspace{0.2em}
\noindent \textbf{Dataset details.} We use the Mozilla Common Voice~\cite{mcv} accented English \dataset~benchmarking dataset%
\footnote{\url{https://tinyurl.com/accent-dataset}}
in all our experiments. This dataset consists of five seen (AUS, CAN, SCT, UK, US) and nine unseen accents (AFR, HKG, IND, IRL, MAL, NWZ, PHL, SGP, WLS). The dataset includes \dataset-100 and \dataset-600 training splits containing $100$ hours and $620$ hours of audio data respectively, along with $17$ hours of validation and test splits. For pretraining, we only utilize the larger \dataset-600 train split and for ASR, we report results on both training splits. 
\looseness=-1

\vspace{0.2em}
\noindent \textbf{Pre-training setup.} In all our experiments, we utilize the HuBERT-\textsc{Base} architecture~\cite{hubert}. This architecture consists of a $7$-layer convolution feature extractor and a $12$-layer Transformer Encoder with $12$ attention ($d=768$) heads. The model is trained for two iterations, each consisting of $200$k steps. In both iterations, we use $500$ hidden units obtained from the output of the $6^{\text{th}}$ encoder layer via K-means clustering as the pseudo targets. 
In each iteration, the model is trained using an Adam optimizer~\cite{adam} with learning rate of $5e\text{-}5$, $32$k warmup steps, a batch size of $87.5$ seconds and gradient accumulation over four steps.

\vspace{0.2em}
\noindent \textbf{Fine-tuning setup.} We use the joint CTC-attention based encoder-decoder architecture~\cite{jointctc}, but instead of the standard encoder, we replace it with the model from the self-supervised pre-training~\ref{subsec:method:pretrain} step. For experiments on the \dataset-100 training split, we add 3-way speed perturbation prior to training. Throughout all our experiments, we train the model for $50$ epochs with a batch size of $128$, dropout rate of $0.1$, a learning rate of $1.0$ and $25K$ warmup steps. The loss is calculated with a CTC weight of 0.3 and label smoothing of 0.1.

\section{Experimental Results and Analysis}

Table~\ref{table:main_res} compares the performance~(WER \%) of our proposed system against four alternative approaches: 
\begin{enumerate*}
    \item Conformer baseline~\cite{conformer}
    \item Replacing the encoder in the Conformer with a pre-trained HuBERT model~\cite{hubert}, pretrained with the SSL objective on data \dataset-600
    \item Jointly training HuBERT baseline with an accent classifier (MTL)~\cite{asr_clf}
    \item Domain Adversarial Training (DAT) of the HuBERT baseline with an accent classifier on the $12^{\text{th}}$ encoder layer.
\end{enumerate*}
\begin{table*}[t!]
\centering
\caption{Comparison of the performances (WER\%) of inferences done in absence of particular accents.}
\label{table:res_oracle_single_inf}
\resizebox{\linewidth}{!}{
    \begin{tabular}{ l | ccccc | ccccccccccc  }
        \hline
        
        \multicolumn{1}{c|}{\multirow{2}{*}{\centering \textbf{\small{Accent used}}}} & \multicolumn{5}{c|}{\textbf{\footnotesize{Seen Accents}}} & \multicolumn{9}{c}{\textbf{\footnotesize{Unseen Accents}}} \\
        
        \cline{2-15}

         & \textbf{\scriptsize{AUS}} & \textbf{\scriptsize{CAN}} & \textbf{\scriptsize{UK}} &
        \textbf{\scriptsize{SCT}}  & \textbf{\scriptsize{US}}  & 
        \textbf{\scriptsize{AFR}} & \textbf{\scriptsize{HKG}} & \textbf{\scriptsize{IND}} & \textbf{\scriptsize{IRL}} & \textbf{\scriptsize{MAL}} & \textbf{\scriptsize{NWZ}} & \textbf{\scriptsize{PHL}} & \textbf{\scriptsize{SGP}} & \textbf{\scriptsize{WLS}}  \\
        
        \hline
        \small{\textsc{Codebook} Attention } &  \cellcolor{gray!10} \footnotesize{ 3.7} &  \cellcolor{gray!10} \footnotesize{ 7.5} &  \cellcolor{gray!10} \footnotesize{ 5.5} &  \cellcolor{gray!10} \footnotesize{ 4.7} &  \cellcolor{gray!10} \footnotesize{ 5.9} &  \cellcolor{gray!10} \footnotesize{ 10.8} &  \cellcolor{gray!10} \footnotesize{ 14.7} &  \cellcolor{gray!10} \footnotesize{ 11.6} &  \cellcolor{gray!10} \footnotesize{ 11.4} &  \cellcolor{gray!10} \footnotesize{ 16.7} &  \cellcolor{gray!10} \footnotesize{ 9.2} &  \cellcolor{gray!10} \footnotesize{ 14.9} &  \cellcolor{gray!10} \footnotesize{ 18.3} &  \cellcolor{gray!10} \footnotesize{ 7.3} \\

        \hline
        
         \small{ \texttt{Australia} } & \cellcolor{red!20} \footnotesize{\textbf{4.9}} & \footnotesize{7.4} & \footnotesize{5.6} & \footnotesize{4.9} & \footnotesize{6.0} & \footnotesize{ 10.9} & \footnotesize{ 14.9} & \footnotesize{ 11.4} & \footnotesize{ 11.3} & \footnotesize{ 17.2} & \cellcolor{red!20} \footnotesize{\textbf{9.9}} & \footnotesize{ 14.8} & \footnotesize{ 18.2} & \footnotesize{ 7.6 } \\
    
        \small{ \texttt{Canada} } & \footnotesize{3.7} & \footnotesize{7.6} & \footnotesize{5.5} & \footnotesize{4.8} & \footnotesize{6.1} & \footnotesize{ 10.7} & \footnotesize{ 14.8} & \footnotesize{ 11.6} & \footnotesize{ 11.5} & \footnotesize{ 16.6} & \footnotesize{ 9.1} & \footnotesize{ 14.8} & \footnotesize{ 18.4} & \footnotesize{ 7.0} \\

        \small{ \texttt{England} } & \footnotesize{4.1} & \footnotesize{7.4} & \cellcolor{red!20} \footnotesize{\textbf{5.7}} & \footnotesize{4.7} & \footnotesize{6.1} & \cellcolor{red!20} \footnotesize{\textbf{11.0}} & \footnotesize{ 14.6} & \footnotesize{ 11.7} & \footnotesize{ 11.5} & \footnotesize{ 16.6} & \footnotesize{ 9.3} & \cellcolor{red!20} \footnotesize{\textbf{15.0}} & \footnotesize{ 18.6} & \cellcolor{red!20} \footnotesize{\textbf{7.7}}\\

        \small{ \texttt{Scotland} } &  \footnotesize{3.6} & \footnotesize{7.4} & \footnotesize{5.5} & \cellcolor{red!20} \footnotesize{\textbf{5.2}} & \footnotesize{5.8} & \footnotesize{ 10.7} & \footnotesize{ 14.3} & \footnotesize{ 11.3} & \footnotesize{ 11.5} & \footnotesize{ 16.5} & \footnotesize{ 9.0} & \footnotesize{ 14.7} & \footnotesize{ 18.6} & \footnotesize{ 7.1} \\

        \small{ \texttt{US} } & \footnotesize{3.8} & \footnotesize{7.7} & \footnotesize{5.6} & \footnotesize{4.4} & \footnotesize{6.1} & \footnotesize{ 10.8} & \footnotesize{ 14.8} & \footnotesize{ 11.9} & \footnotesize{ 11.5} & \footnotesize{ 17.2} & \footnotesize{ 9.2} & \footnotesize{ 14.9} & \footnotesize{ 18.5} & \footnotesize{ 7.3} \\

        \small{ \texttt{US + Canada} } & \footnotesize{3.6} &  \cellcolor{red!20} \footnotesize{\textbf{8.0}} & \footnotesize{5.6} & \footnotesize{4.8} & \cellcolor{red!20} \footnotesize{\textbf{6.4}} & \footnotesize{ 10.7} & \cellcolor{red!20} \footnotesize{\textbf{15.4}} & \cellcolor{red!20} \footnotesize{\textbf{12.3}} & \cellcolor{red!20} \footnotesize{\textbf{11.9}} & \cellcolor{red!20} \cellcolor{red!20} \footnotesize{\textbf{17.3}} & \footnotesize{ 9.1} & \footnotesize{ 14.9 } & \cellcolor{red!20} \footnotesize{\textbf{18.9}} & \footnotesize{ 7.0} \\

        \hline
    \end{tabular}
}
\end{table*}
Similar to previous studies~\cite{baevski2020wav2vec,hubert}, we find that replacing Conformer's encoder with a pre-trained HuBERT model leads to significant improvement in performance; we note here that the HuBERT-based encoder is pretrained from scratch on \dataset-600. Furthermore, we observe that, during self-supervised pre-training, using an existing checkpoint (i.e., the HUBERT-\textsc{Base} Librispeech checkpoint from Fairseq~\cite{fairseq}) leads to additional performance gains. 
Given the limited amount of supervised finetuning data in \dataset-100, to combat overfitting, we freeze the feature extractor and the first $3$ layers of the HuBERT encoder; this yields further benefits. (In all subsequent experiments, we will use this best setup of selective ASR finetuning.) We also compare our system against the MTL and DAT approaches. Overall, our proposed approach performs significantly better on nearly all accents, particularly the unseen accents. 

\begin{table}[t!]
\centering
\caption{Comparing zero-shot performance~(WER \%) of our architecture with other approaches on the L2Arctic dataset. $\dagger$ indicates statistical significance (at $p$ $<$0.001 using MAPSSWE test) w.r.t. the HuBERT baseline.}
\label{table:res_l2arctic}
\resizebox{\linewidth}{!}{
\begin{tabular}{ | l | c | cccccc | }
    \hline
    \multicolumn{1}{|c|}{\multirow{2}{*}{\textbf{\small{Method}}}} & \multirow{2}{*}{\textbf{\small{All}}} & \multicolumn{6}{c|}{\textbf{\footnotesize{Accents}}} \\
    
    \cline{3-8}
    
    &  &  \textbf{\scriptsize{ARA}} & \textbf{\scriptsize{HIN}} & \textbf{\scriptsize{KOR}} &
    \textbf{\scriptsize{MAN}}  & \textbf{\scriptsize{SPA}}  & 
    \textbf{\scriptsize{VIA}}  \\

    \hline

   \small{HuBERT Encoder}~\cite{hubert} & \footnotesize{ 22.6 } & \footnotesize{ 20.2 } & \footnotesize{ 17.8 } & \footnotesize{ 17.3 } & \footnotesize{ 25.8 } & \footnotesize{ 20.4 } & \footnotesize{ 33.7 }  \\

  \small{MTL}~\cite{asr_clf} & \footnotesize{ 23.0 } &  \footnotesize{ 21.0 } & \footnotesize{ 18.1 } & \footnotesize{ 17.6 } & \footnotesize{ 26.4 } & \footnotesize{ 20.9 } & \footnotesize{ 34.1 } \\

   \small{DAT}~\cite{bobw} & \footnotesize{ 22.9 } &  \footnotesize{ 20.7 } & \footnotesize{ 18.2 } & \footnotesize{ 17.4 } & \footnotesize{ 26.2 } & \footnotesize{ 20.9 } & \footnotesize{ 34.1 } \\

   \hline

   \small{\textsc{Codebook} Attention} & \footnotesize{\textbf{ 21.7 }$\dagger$} &  \footnotesize{\textbf{ 19.9 }} & \footnotesize{\textbf{ 16.5 }} & \footnotesize{\textbf{ 16.4 }} & \footnotesize{\textbf{ 24.8 }} & \footnotesize{\textbf{ 19.8 }} & \footnotesize{\textbf{ 32.7 }} \\
    
    \hline
\end{tabular}
}
\vspace{-0.5cm}
\end{table}

\vspace{0.3em}
\noindent \textbf{Importance of Codebooks.} Table~\ref{table:res_oracle_single_inf} shows the WERs of our best codebook-based system when the codebook of the seen accent is withheld during inference with the joint beam search. The first row shows the WERs when all codebooks are present. This is to check whether the absence of the codebook of the seen accent during decoding degrades performance. On removing the Australian codebook, we find that the WER worsens from $3.7\%$ to $4.9\%$. Interestingly, WER of the unseen New-Zealand accent suffers the most when the Australian codebook is withheld ($9.2 \rightarrow 9.9$). This suggests that relatedness in accents is being captured in the codebooks. Also, on removing both US and Canada accent codebooks (that are closely related), we see the largest degradation in WER on US and Canada test samples.

\vspace{0.3em}
\noindent \textbf{Zero-shot ASR Evaluation.}
In Table~\ref{table:res_l2arctic}, we compare WERs of our proposed system with baselines when evaluated in a zero-shot setting on out-of-domain accented samples from the L2-Arctic~\cite{l2arctic} dataset. The dataset consists of six non-native English accents namely: Arabic~(ARA), Hindi~(HIN), Korean~(KOR), Mandarin~(MAN), Spanish~(SPA), and Vietnamese~(VIA). Our system significantly outperforms the baselines (at $p<0.001$) across all accents on the L2-Arctic dataset. This attests to the robustness of the learned speech representations from our system.

\vspace{0.3em}
\noindent \textbf{Ablation Analysis.} In Table~\ref{table:res_ablation}, we investigate various settings related to the codebooks. We apply codebooks to different encoder layers and obtain the best performance when applying codebooks on the $6^{\text{th}}$ layer for seen accents, and on all 12 layers for unseen accents. With varying codebook sizes, $50$ codebook entries performed the best on both seen/unseen accents; 500 codebook entries improved the seen accents further but hurt the unseen accents, thus indicating overfitting. Lastly, we examined different training protocols for the codebooks. Randomly-initialized (and not learnable) codebooks ($C_\text{random}$) do nearly as well as codebooks trained during both SSL and ASR fine-tuning (denoted by $C_\text{both}$). We also evaluate with frozen codebooks that are learned during pretraining and frozen during ASR fine-tuning ($C_\text{frozen}$). This performs nearly as well as $C_\text{both}$ on seen accents but slightly underperforms on unseen accents. This suggests that the codebooks are meaningfully learned during the pretraining stage.


\renewcommand*{\arraystretch}{1.05}
\begin{table}[t!]
\centering
\caption{Comparison of the performance (WER \%) of different variants of our architecture.
$\textsc{C}_{L \in (i,\ldots,j)}(P=k)$: Codebook attention applied at all layers from $i$ to $j$ with $k$ entries per accent codebook. }
\label{table:res_ablation}
\resizebox{\linewidth}{!}{
\begin{tabular}{ | l | c | c | c | c | }
    \hline
    
    \multicolumn{2}{|c|}{\textbf{\small{Setup}}} & \multirow{2}{*}{\textbf{\small{Overall}}} & \multirow{2}{*}{\textbf{\small{Seen}}} & \multirow{2}{*}{\textbf{\small{Unseen}}} \\
    
    \cline{1-2}
    
    \multicolumn{1}{|c|}{\vspace{-0.17cm}} & \textbf{\vspace{-0.17cm}} & & & \\
    \multicolumn{1}{|c|}{\textbf{\small{Layers}}} & \textbf{\small{\# codebooks}} & & & \\
    
    \hline
    \multicolumn{5}{|c|}{\vspace{-0.4cm}} \\[0.03cm]
    \multicolumn{5}{|c|}{\textbf{\footnotesize{Varying codebook influence}}} \\[0.03cm]
    \hline

    \small{$L = \{ 6 \} $} & \small{$C=50$} & \footnotesize{9.01} &  \footnotesize{\textbf{5.89}} & \footnotesize{12.12} \\

    \small{$L = \{ 1,\ldots,6 \} $} & \small{$C=50$} & \footnotesize{9.03} &  \footnotesize{5.91} & \footnotesize{12.13} \\
    
    \small{$L = \{1,\ldots,12 \}$} & \small{$C=50$} & \footnotesize{\textbf{8.95}} &  \footnotesize{5.94} & \footnotesize{\textbf{11.91}} \\
    
    \hline
    \multicolumn{5}{|c|}{\vspace{-0.4cm}} \\[0.03cm]
    \multicolumn{5}{|c|}{\textbf{\footnotesize{Varying codebook size}}} \\[0.03cm]
    \hline

    \small{$L = \{1,\ldots,12 \}$} & \small{$C=50$} & \footnotesize{\textbf{8.95}} &  \footnotesize{5.94} & \footnotesize{\textbf{11.91}} \\

    \small{$L = \{1,\ldots,12 \}$} & \small{$C=200$} & \footnotesize{9.01} &  \footnotesize{6.00} & \footnotesize{12.10} \\

    \small{$L = \{1,\ldots,12 \}$} & \small{$C=500$} & \footnotesize{9.03} &  \footnotesize{\textbf{5.87}} & \footnotesize{12.18} \\

    \hline 
    \multicolumn{5}{|c|}{\vspace{-0.4cm}} \\[0.03cm]
    \multicolumn{5}{|c|}{\textbf{\footnotesize{Varying codebook nature}}} \\[0.03cm]
    \hline

    \multicolumn{1}{|c|}{\small{$L = \{ 6 \}$}} & \small{$C_{\text{both}}=50$} & \footnotesize{9.01} &  \footnotesize{\textbf{5.89}} & \footnotesize{12.12} \\

    \multicolumn{1}{|c|}{\small{$L = \{ 6 \}$}} & \small{$C_{\text{frozen}}=50$} & \footnotesize{9.07} &  \footnotesize{5.91} & \footnotesize{12.23} \\

    \multicolumn{1}{|c|}{\small{$L = \{ 6 \}$}} & \small{$C_{\text{random}}=50$} & \footnotesize{\textbf{8.98}} &  \footnotesize{5.93} & \footnotesize{\textbf{12.02}} \\
    
    \hline

\end{tabular}
}

\vspace{-0.15in}
\end{table}
\renewcommand*{\arraystretch}{1}

\vspace{0.3em}
\noindent \textbf{ASR Fine-tuning with Larger Dataset.} 
We fine-tune the pretrained HuBERT baseline, MTL, DAT, and our best system on \dataset-600. As seen in Table~\ref{table:res_bigger_dataset}, our system performs comparably to the other baselines on the seen accents but significantly improves on unseen accents.

\begin{table}[t]
\centering
\caption{Comparison of WERs\% of our approach compared to other baselines on \dataset-600 dataset.}
\label{table:res_bigger_dataset}
\resizebox{\linewidth}{!}{
\begin{tabular}{ | l | c | c | c | }
    \hline

    \multicolumn{1}{|c|}{\textbf{\small{Method}}} & \textbf{\small{Overall}} & \textbf{\small{Seen}} & \textbf{\small{Unseen}} \\

    \hline

   \small{HuBERT~\cite{hubert}} & \footnotesize{6.68} &  \footnotesize{3.87} & \footnotesize{9.49} \\

   \small{MTL~\cite{asr_clf}} & \footnotesize{6.57} &  \footnotesize{\textbf{3.76}} & \footnotesize{9.37} \\

  \small{DAT~\cite{bobw}} & \footnotesize{6.57} &  \footnotesize{3.83} & \footnotesize{9.30} \\

    \hline

    \small{\textsc{Codebook} Attention } & \footnotesize{ \textbf{6.43} } &  \footnotesize{ 3.80 } & \footnotesize{ \textbf{9.19} } \\

   \hline
    
\end{tabular}
}
\vspace{-0.5cm}
\end{table}

\section{Conclusion}
In this work, we propose an accent-aware ASR adaptation technique where accent-specific codebooks are incorporated within the Transformer layers of a HuBERT model via cross-attention. This integration happens right from the SSL-based pretraining stage. The pretrained codebooks and encoder layers are further finetuned using supervised ASR fine-tuning. Compared to existing accent adaptation techniques, we observe that this yields significant WER reductions on English utterances in both seen and unseen accents in the Mozilla Common Voice (MCV) corpus. The accent-aware models trained on MCV also generalize well to out-of-domain accented English samples (from a different corpus, L2Arctic) when evaluated in a zero-shot setting. In future work, we aim to use self-training with unlabeled data (with accent labels) to further refine the accent codebooks.

\section{Acknowledgements}
We acknowledge the financial support from a SERB Core Research Grant, Department of Science and Technology, Government of India on accented speech processing.

\section{References}
\printbibliography

\end{document}